\documentclass[letterpaper, 10 pt, conference]{ieeeconf}  

\IEEEoverridecommandlockouts                              

\usepackage{graphics}
\usepackage{multirow}

\usepackage{graphics} 
\usepackage{epsfig} 
\usepackage{times} 
\usepackage{amsmath} 
\usepackage{amssymb}  
\usepackage{wasysym}
\PassOptionsToPackage{hidelinks}{hyperref}
\usepackage{array}
\usepackage{booktabs}
\usepackage{siunitx}
\usepackage{algorithm}
\usepackage{algpseudocode}

\usepackage[hidelinks]{hyperref}
\usepackage{caption}
\usepackage[T1]{fontenc}  

\usepackage{xcolor,pifont}
\usepackage{color}
\usepackage{acronym}
\usepackage{cite}
\usepackage{tabularx} 
\usepackage{multirow}
\usepackage{array}

\usepackage{cleveref}

\title{\LARGE \bf

ORCA: An Open-Source, Reliable, Cost-Effective, Anthropomorphic Robotic Hand for Uninterrupted Dexterous Task Learning}

\author{
\begin{minipage}[t]{0.9\textwidth}
\centering 
 Clemens C. Christoph$^{1}$\textsuperscript{†}, 
 Maximilian Eberlein$^{1}$\textsuperscript{†},
 Filippos Katsimalis$^{1}$\textsuperscript{†},
 Arturo Roberti$^{1}$\textsuperscript{†},
 Aristotelis Sympetheros$^{1}$\textsuperscript{†},
 Michel R. Vogt$^{1}$\textsuperscript{†},
 Davide Liconti$^{1}$,
 Chenyu Yang$^{1}$,
 Barnabas Gavin Cangan$^{1}$,
 Ronan J. Hinchet$^{1}$,
 Robert K. Katzschmann$^{1*}$
\end{minipage}
\thanks{$^{1}$Soft Robotics Lab, IRIS, D-MAVT, ETH Zurich, Switzerland
        }
\thanks{† Equal contribution.}
\thanks{$^{*}$Corresponding author: \href{mailto:rkk@ethz.ch}{\tt rkk@ethz.ch}}
}

\begin{document}

\maketitle

\begin{abstract}
General-purpose robots should possess human-like dexterity and agility to perform tasks with the same versatility as us. A human-like form factor further enables the use of vast datasets of human-hand interactions. However, the primary bottleneck in dexterous manipulation lies not only in software but arguably even more in hardware. Robotic hands that approach human capabilities are often prohibitively expensive, bulky, or require enterprise-level maintenance, limiting their accessibility for broader research and practical applications. What if the research community could get started with reliable dexterous hands within a day? We present the open-source ORCA hand, a reliable and anthropomorphic 17-DoF tendon-driven robotic hand with integrated tactile sensors, fully assembled in less than eight hours and built for a material cost below 2,000 CHF. We showcase ORCA's key design features such as popping joints, auto-calibration, and tensioning systems that significantly reduce complexity while increasing reliability, accuracy, and robustness. We benchmark the ORCA hand across a variety of tasks, ranging from teleoperation and imitation learning to zero-shot sim-to-real reinforcement learning. 
Furthermore, we demonstrate its durability, withstanding more than 10,000 continuous operation cycles---equivalent to approximately 20 hours---without hardware failure, the only constraint being the duration of the experiment itself.
Video is here: \texttt{\href{https://youtu.be/kUbPSYMmOds}{youtu.be/kUbPSYMmOds}}.  Design files, source code, and documentation are available at
\texttt{\href{https://srl.ethz.ch/orcahand}{srl.ethz.ch/orcahand}}.
\end{abstract}

\section{Introduction}\label{introduction}
Reproducing the intricate dexterity of the human hand has long been a central challenge in robotics \cite{grasping2000, dexmanip2000}. Although robotic grippers excel in industrial automation, their limited versatility makes them unsuitable for interaction with tools and objects designed for human hands \cite{manipdex2022, grippers2023}. Consequently, extensive research has been devoted to developing anthropomorphic robotic hands and training them to solve complex manipulation tasks \cite{review2023, HUANG2025100212}. However, compared to grippers, anthropomorphic hands require significantly more actuators, increasing the complexity of their assembly and control. In addition, good anthropomorphic hand hardware must be durable, repeatable, and versatile to be used in machine learning applications. Whether it is the sim2real gap in reinforcement learning (RL) or the accuracy of teleoperation in imitation learning (IL), bottlenecks in dexterous manipulation stem not only from software, but also, and maybe more importantly, from hardware limitations \cite{metric2013,challenges2019,zhu2022challenges}.

\begin{figure}[t!]
\centering   
\includegraphics[width=\linewidth]{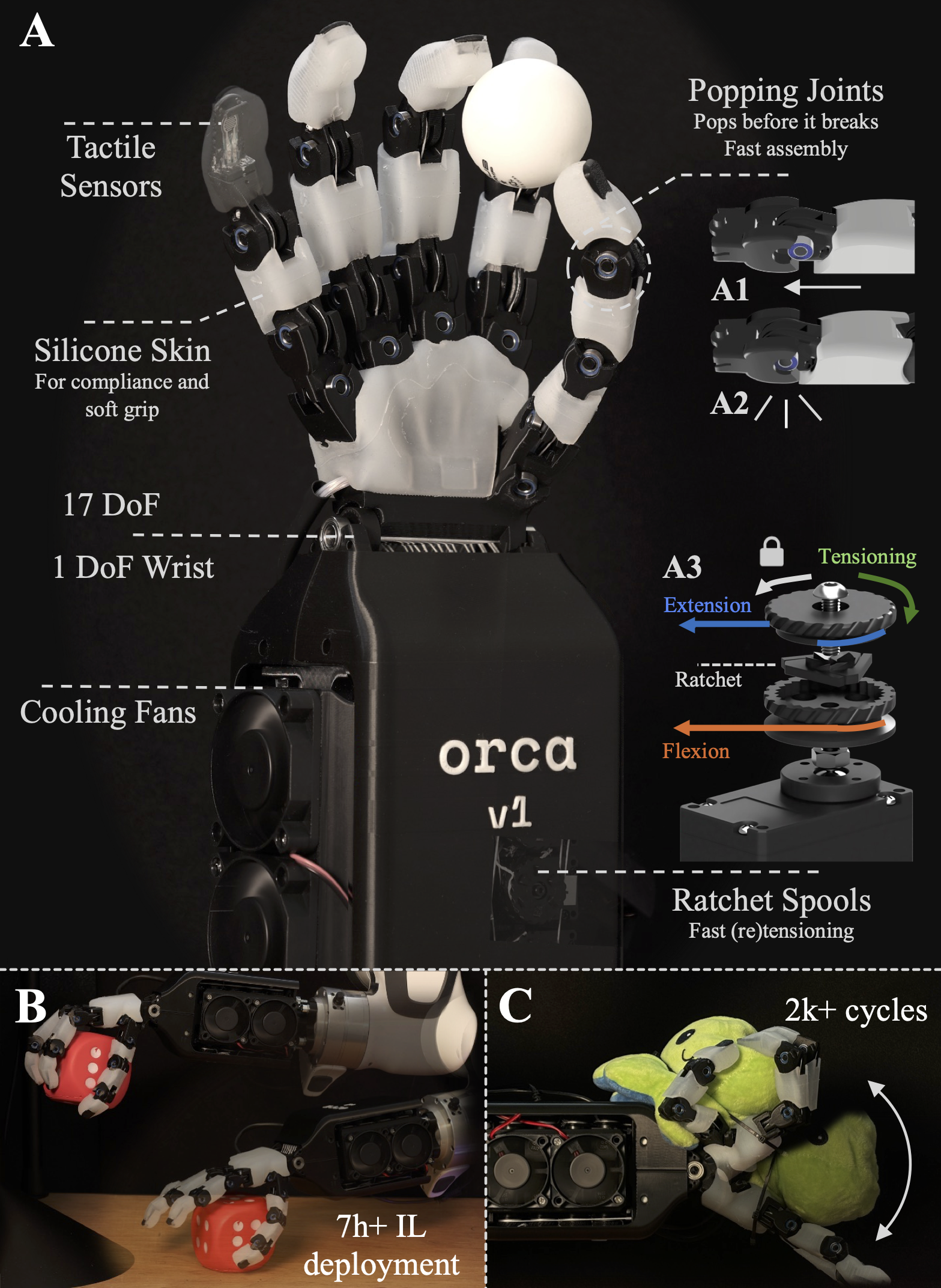   }
    \caption{(A) The ORCA hand closely mimics its human counterpart with the same form factor, a bony structure, and silicone-cast skin. The ORCA hand is 3D-printed but incorporates joints designed to pop before breaking, making it resistant to overload-induced failures while retaining the advantages of bearing pinhole joints, such as stability and simple kinematics. (A1) Just before the joint pops. (A2) Applying pressure pops the joint into place and keeps it secure. (A3) Depicts our spool system, which enables manual retention without unscrewing the spools or tendons. (B) We show that our hand can be deployed in real-world settings by running our self-resetting imitation learning policy for over 7 hours before we decided to end the experiment. (C) Our reliability test reveals our hand's robustness and the high repeatability of joint movements.
}
\label{fig:1}
\vspace{-15pt}
\end{figure}

\begin{figure*}[t!]
\vspace*{0.2cm}
\centering   
\includegraphics[width=\linewidth]{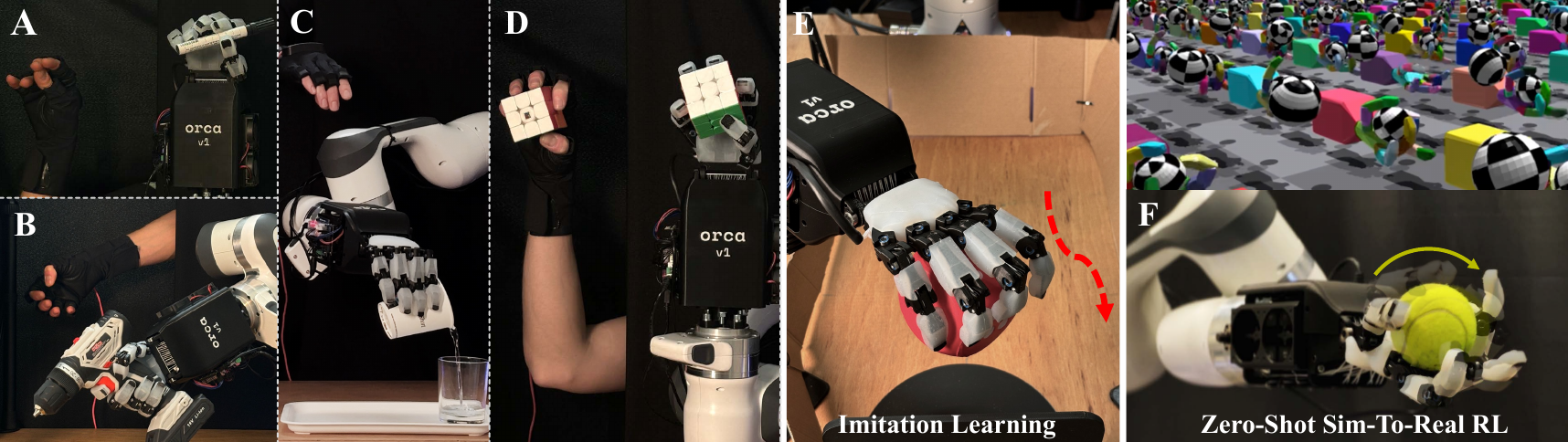}
\caption{Versatility of the ORCA hand: (A)-(D) Teleoperation with ROKOKO \cite{rokoko_website}
 gloves. (A) Holding a pen (B) Using a drill, showing high dexterity. (C)  Liquid pouring. (D) Grasping a cube: This picture illustrates how closely the ORCA hand resembles a human hand. (E) IL with walls and a slider for self-resetting.
(F) Policies in simulation, such as rolling a ball, can be deployed zero-shot to the real world due to the ORCA hand's low joint errors. }
\label{fig:versatility}
\vspace{-15pt}
\end{figure*}
Tendon-driven robotic hands have been a focal point of robotics research since its early days \cite{bekey1990control,salisbury1982articulated,jacobsen1986design}. Among the various designs, the proprietary Shadow Hand \cite{shadow_robot_dexterous_hand} demonstrated impressive capabilities in dexterous manipulation tasks \cite{in_hand_manipulation_2020}. However, these hands cost over 100,000 CHF, require substantial maintenance \cite{openai2019solvingrubikscuberobot} and are difficult to repair due to their proprietary and highly integrated designs. Other recent tendon-driven hand designs, such as the InMoov hand \cite{inmoov_hand} and the DexHand \cite{dexhand}, offer the advantages of being open source and low-cost. However, the InMoov hand is limited in dexterity, while the DexHand is challenging to assemble, and neither has demonstrated real-world applicability in autonomous manipulation tasks. 

Alternatives to tendon-driven hands are direct-driven hands, such as the Allegro Hand \cite{allegro_hand}, priced at around 15,000 CHF, and the open-source LEAP hand \cite{shaw2023leaphandlowcostefficient}, which is even more cost-effective under 2,000 CHF and requires only three hours of assembly for unprecedented levels of reliability. However, all direct-driven hand designs share limitations such as bulkiness, restricted form factors, and an inability to match the softness, form factor, and agility of a human hand. Embedding motors within the fingers increases inertia and limits power output, restricting the quick, dynamic, and forceful movements essential for human-like motion.

In this paper, we present the ORCA hand: a tendon-driven, dexterous, and anthropomorphic robotic hand with fully integrated tactile sensors. The ORCA hand is designed for reliability, simplicity, and versatility in a wide range of tasks. Key contributions of our integrated system include:
\begin{itemize} 
\item An open source, 3D-printable design with a cost of less than 2,000 CHF, which can be assembled by a single person without prior experience in less than eight hours. 
\item A joint design that pops before it breaks, enhancing the durability of 3D-printed components and streamlining the assembly process.
\item Auto-calibration enabled by tendon routing through the center of rotation. This minimizes joint position errors and increases repeatability.
\item Fully integrated tactile sensors and sensor wiring, which can be produced in-house, offering a compact and modular solution.
\end{itemize}

We demonstrate the hand's dexterity by teleoperating it to perform complex tasks that traditional robotic grippers cannot accomplish. Through a variety of reliability tests we demonstrate the ORCA hand's capability to offer exceptional reliability, durability, and consistent performance during tens of hours of operation. To showcase the ORCA hand's dexterity and accuracy we implement fine motor control tasks like in-hand object orientation. Leveraging the anthropomorphic design, we are also able to implement imitation learning tasks like the picking and placement of cubes and execute these autonomous tasks continuously for multiple hours without any human intervention on the hand hardware.


\section{System Design}\label{systemDesign}
The ORCA hand design follows the requirements set above, namely dexterity and reliability at minimal complexity and cost. It comprises five fingers, including an opposable thumb and an actuated wrist (\Cref{fig:1}-A), and is of a size similar to the average human hand \cite{hand_anatomy}. The total weight of the hand is approximately 1.2 kg. The fingers are mounted on a base that resembles the human palm containing the carpal and metacarpal bones. The palm is connected to the wrist mechanism that is mounted on the \textit{tower}. The tower contains the motors and all auxiliary electronics, and is enclosed in a protective casing. The rest of this section describes the most important design features of the ORCA hand in detail.

\subsection{Tendon Actuation for Agility}
Each joint, except for the wrist joint, is actuated using two fishing lines (Nylon fibers braided into a 0.4 mm diameter rope) under tension, here referred to as tendons. One tendon is responsible for flexion (flexor) and the other for extension (extensor). This decision was made based on the observation that a smaller form factor and lower finger inertia more closely mimic the nimbleness and dexterity of human hands compared to direct-driven hands. In addition, tendon actuation makes the ORCA hand independent of the choice of actuator, enabling actuation technologies other than electric motors to be used in the future, such as contracting artificial muscles.

\begin{figure}[t!]
\vspace*{0.2cm}

\centering
\includegraphics[width=\linewidth]{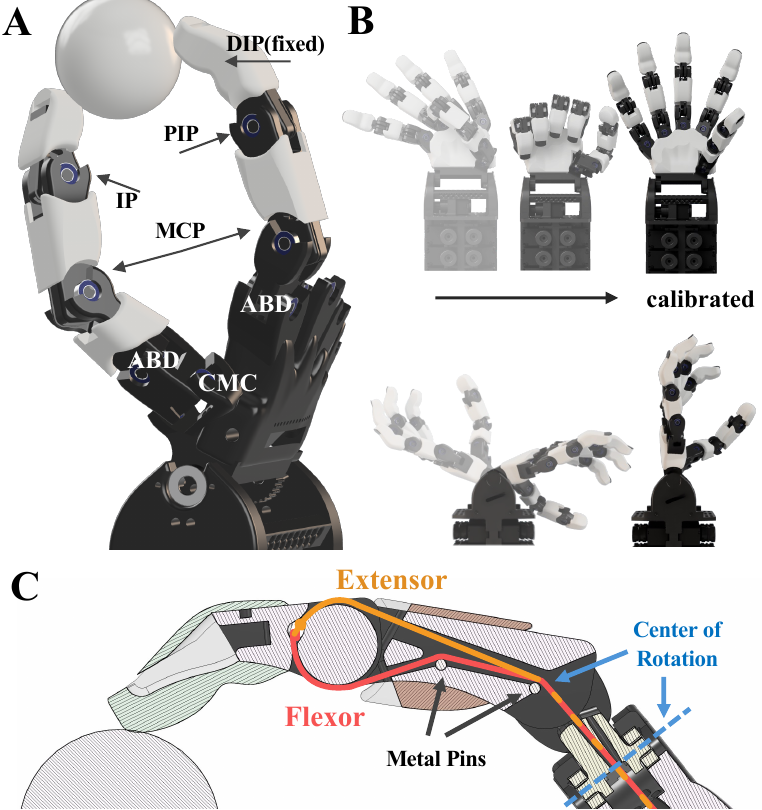}
\caption{ (A) Naming convention of the joints. The thumb includes an additional degree of freedom. (B) Auto-calibration: The three-step process moves all joints to their respective limits and determines a mapping between motor and joint angles without any external sensors.
(C) Routing of the DIP joint. The tendons are guided around metal pins, to reduce friction and eliminate wear over time. Moreover, they are always guided through the center of rotation for straightforward control at minimal slack. }
\label{fig:routing}
\vspace{-15pt}
\end{figure}

Although tendon actuation offers many advantages, it also presents challenges such as friction build-up, wear, and slack over time, which can affect movement precision and longevity. We mitigate those challenges as follows:
\begin{itemize}
    \item We avoid direct contact of the tendons with polylactic acid (PLA) by deflecting the tendons around smooth metal pins and rods, as depicted in  \Cref{fig:routing}-C. 
    \item We use Teflon tubes for nonlinear routing, \textit{e.g.}, from the bottom of the thumb to the wrist.
    \item Finally, tendons can be manually re-tensioned using a ratchet spool mechanism mounted on the motors, as shown in \Cref{fig:1}-A3. The ratchet is attached to the top spool, allowing rotation in one direction while locking movement in the other. This design makes the ORCA hand user-friendly, as re-tensioning can be done in seconds without the need to unscrew the spool or tendon. The spool system quickly removes any slack that accumulates. In addition, tendons can be easily loosened, allowing joints to pop out for quick replacement of broken parts.
\end{itemize}


\subsection{Poppable Pin Joints}

Rolling contact joints \cite{rolling-contact-srl,dexhand} have become a popular alternative to pinhole joints for 3D-printed hands due to their ability to dislocate instead of break. However, these mechanisms require ligaments that can potentially loosen over time and increase complexity. We introduce a pin joint design that allows the joints to "pop" out of place and dislocate instead of breaking when excessive radial and axial loads are applied (\Cref{fig:1}-A1). 

This feature is achieved by placing the bearings in circular arc-shaped grooves, which hold them tightly under normal operating conditions, but allow them to dislocate in the event of a forceful collision. This mechanism combines the advantages of pinhole joints, such as axial stability and straightforward kinematics, with the robustness of ligament-based rolling contact joints, while being extremely quick and easy to assemble.

\subsection{Finger and Palm Design}
Fingers two to five (index to pinky) have three actuated joints that mimic those of the human finger: the Proximal Interphalangeal (PIP), the Metacarpophalangeal (MCP), and the Abduction (ABD) joint (\Cref{fig:routing}-A). The ORCA PIP joint corresponds to the human PIP joint, while the MCP and ABD joints together correspond to the human MCP joint which can perform both flexion/extension and abduction/adduction, which is necessary in various dexterous manipulation tasks \cite{abduction_importance}. The range of motion (RoM) for each joint is shown in \Cref{tab:table_ROM}, based on human anatomy \cite{hand_anatomy}.

\begin{table}[h!]
\centering
\begin{tabularx}{\linewidth}{lXXXX}
\toprule
\multirow{2}{*}{\textbf{Joint Name}} & \multicolumn{2}{c}{\textbf{Fingers 2 to 5}} & \multicolumn{2}{c}{\textbf{Thumb}} \\
\cmidrule(lr){2-3} \cmidrule(lr){4-5}
        
 & \textbf{Flexion} & \textbf{Extension} & \textbf{Flexion} & \textbf{Extension} \\ 
 \midrule
\textbf{IP} & - & - & 100° & 20° \\ 
\textbf{PIP} & 130° & 20° & - & - \\ 
\textbf{MCP} & 110° & 20° & 115° & 20° \\ 
\textbf{ABD} & 30° & 30° & 45° & 45° \\ 
\textbf{CMC} & - & - & 48° & 53° \\ \bottomrule
\end{tabularx}
\caption{Motion range of the joints of the ORCA hand.}
\label{tab:table_ROM}
\end{table}

The removal of the DIP joints on digits 2 to 5 was an intentional design decision. In most cases, DIP joints are rigidly coupled with the corresponding PIP joints and do not move independently, meaning they do not contribute additional active degrees of freedom. By fixing the DIP joints, we eliminate the potential for slack buildup over time and enable direct motor control of the distal joint, resulting in improved accuracy and predictability. Furthermore, this approach simplifies assembly significantly and frees up space for additional sensing electronics.
The DIP joint is not directly actuated in the human hand, and as such it is of less importance compared to the other hand joints in dexterous manipulation tasks. A positive side-effect of the removal of the DIP joint is the substantial increase in the space available for tactile sensing integration on the fingertips.

The thumb differs from other fingers and has four instead of three DoFs, \textit{i.e.}, the Interphalangeal (IP), MCP, ABD, and Carpometacarpal (CMC) joints. Additionally, the thumb is positioned with a supination of $15^\circ$ on the palm, making it opposable to the other fingers \cite{hand_anatomy}.

\subsection{Wrist Design}

A major limitation of hands without a wrist joint is their inability to orient the palm parallel to surfaces like a table, as the robotic arm obstructs movement. This lack of a wrist joint significantly impairs grasping performance. Motivated by this, we added one rotational DoF around the transverse (radioulnar) axis of the hand. The human wrist can flex and extend about $80^\circ$ \cite{hand_anatomy}, while the ORCA wrist mechanism is capable of achieving $60^\circ$ in flexion and extension.
Instead of using tendon actuation for the wrist, we opted for a belt drive to account for the increased loads the wrist experiences. We use a standard GT2 timing belt with fiberglass reinforcement, which exhibits negligible slack buildup over time.
We decided not to add a second DoF at the wrist to represent the radial/ulnar deviation of the human wrist, particularly to avoid unnecessary complexity and increased cost, since the human wrist is significantly limited in this type of motion compared to flexion/extension \cite{hand_anatomy}.



\subsection{Integrated tactile sensing}

Many previous works have enhanced the manipulation abilities of robotic hands by integrating tactile sensors, particularly on the fingertips. These sensors include force sensors \cite{yin2023rotatingseeinginhanddexterity, multimodalsensor2020}, piezoresistive pressure sensors \cite{egli2024sensorizedsoftskindexterous}, capacitive pressure sensors \cite{tactilesensor2010}, and Hall effect sensors \cite{omnisensor2021}. 

For the ORCA hand, we utilize force sensing resistors (FSR) (RP-C7.6-ST Thin-Film Pressure Sensor) mounted onto a solid FDM-printed PLA backplate, and covered by silicone-molded skin, to provide the hand with binary tactile feedback on all five fingertips. While FSR sensors can theoretically measure the magnitude of applied force, a binary interpretation was chosen due to the compliant skin and its irregular surface, which dampens external forces to varying degrees depending on the contact point on the fingertip. As such, the force magnitude cannot be estimated without additional information about the location of indentation, which the FSR sensors themselves are unable to provide. 

All sensors are connected to external electronics through thin copper wires (\(\varnothing\) 0.2mm), which are routed through the internal finger and palm structure through PTFE tubing, to protect them from the environment and external forces, while also providing a clean visual appearance. To read the output of the FSR sensors, each sensor forms a voltage divider with a \(10 k\Omega\) resistor, the three nodes of which are connected to a 5V input, an analog input, and ground, respectively, all provided by a small microcontroller (Arduino Nano Every).

\subsection{Self-Calibration for Accurate Control}

Consistency across runs is essential in imitation learning, as variability between teleoperation and execution degrades policy performance \cite{belkhale2023dataqualityimitationlearning}, especially in sim-to-real settings. While proprioceptive sensors offer reliable control \cite{shadow_robot_dexterous_hand}, they add cost and complexity. We propose a simpler, cost-effective alternative: automated self-calibration to estimate motor-to-joint mappings (Algorithm 1).

\begin{algorithm}[h!]
\caption{Self-Calibration Algorithm}
\label{alg:self_calibration}
\begin{algorithmic}[1]
\State \textbf{Input:} Set of joints $\mathcal{J} = \{1, \dots, N\}$, joint ROMs $[\theta_j^{\min}, \theta_j^{\max}]$ for each joint $j$
\State \textbf{Output:} Motor limits $[m_j^{\min}, m_j^{\max}]$, motor-to-joint ratios $\rho_j$ for each joint $j$

\For{each joint $j \in \mathcal{J}$}
    \State Move joint $j$ towards its flexion limit until stall is detected
    \State $m_j^{\min} \gets \text{current motor position}$
    \State
    \State Move joint $j$ towards its extension limit until stall is detected
    \State $m_j^{\max} \gets \text{current motor position}$
    \State
    \State Compute motor travel: $\Delta m_j = m_j^{\max} - m_j^{\min}$
    \State Compute joint ROM: $\Delta \theta_j = \theta_j^{\max} - \theta_j^{\min}$
    \State Compute motor-to-joint ratio: $\rho_j = \frac{\Delta m_j}{\Delta \theta_j}$
\EndFor
\end{algorithmic}
\end{algorithm}

Let $\theta_j$ be the true angle and $m_j$ the position of the motor for joint $j$. The ideal goal is to perfectly control $\theta_j$ by commanding $m_j$. However, in practice, especially with tendon-driven hands without joint sensors, there is a discrepancy. A simple model would be $\theta_j = f(m_1, \dots, m_{17}) + \epsilon$, where $f$ is a nonlinear function and $\epsilon$ represents model errors arising from factors like tendon and pulley radii ($r_t, r_p$), tendon slack $s$, servo drift $d$, and manual measurement errors $m_{err}$. These factors can introduce substantial offsets.

Referring to the routing of the ORCA hand, as depicted in \Cref{fig:routing}, each tendon passes through or near the center of rotation (CoR). This design ensures that joint positions are approximately decoupled and can be actuated linearly and independently. This linearity allows us to simplify the model significantly. The self-calibration procedure detailed in \Cref{alg:self_calibration} leverages this by empirically finding the linear relationship for each joint. The process works as follows:

\begin{enumerate}
    \item \textbf{Determine Joint ROM (Line 11):} The absolute range of motion (ROM) for each joint, $\Delta \theta_j$, is precisely determined from the robot's CAD models. This provides an accurate ground truth for the joint's angular travel between its physical stops, $[\theta_j^{\min}, \theta_j^{\max}]$.
    
    \item \textbf{Find Motor Limits (Lines 4-8):} During calibration, each joint $j$ is programmatically moved to its extreme mechanical limits (e.g., fully flexed and fully extended). The corresponding motor positions, $m_j^{\min}$ and $m_j^{\max}$, are recorded when the joint stalls.
    
    \item \textbf{Calculate Transmission Ratio (Lines 10-12):} The total motor travel, $\Delta m_j$, is computed from the recorded limits. The motor-to-joint transmission ratio $\rho_j$ is then calculated via linear interpolation as the ratio of the motor travel to the joint's known ROM:
    $$
    \rho_j = \frac{\Delta m_j}{\Delta \theta_j} = \frac{m_j^{\max} - m_j^{\min}}{\theta_j^{\max} - \theta_j^{\min}}
    $$
\end{enumerate}

This autocalibration largely eliminates the complex error terms related to physical modeling and manual measurements, $\epsilon(r_t, r_p, s, d, m_{err})$, leaving a much smaller residual error, $\epsilon(\Delta \theta_j, s, d)$. With the calibrated parameters, the motor position $m_j$ required to achieve a desired joint angle $\theta_j$ is computed using a linear mapping:
$$
m_j = m_j^{\min} + \rho_j \cdot (\theta_j - \theta_j^{\min})
$$
This method ensures a consistent and accurate mapping from desired joint angles to motor commands across different operational sessions.

\section{Hardware Performance Tests}
\subsection{Reliability and Robustness}\label{sec:reliability-test}

To evaluate the reliability and robustness of the ORCA hand in long-duration tasks, we conducted an experiment in which we actuate the hand joints continuously for 2.5 hours (\Cref{fig:1}-C). We attach a plush animal to the palm of the hand and have it grasp it with all fingers every four seconds. This setup, using a compliant object, is similar to the repeatability test performed by \cite{shaw2023leaphandlowcostefficient} and allows us to assess behavior under increased stress over a greater range of motion. Moreover, to test the durability of the wrist joint, we flex and extend the wrist to $40^\circ$ at one fourth of the frequency of the finger, that is, every 16 seconds. 

The hand successfully completed 2,250 grasping cycles without breakage, motor shutdown, or excessive tendon slack. \Cref{fig:accuracy}-C shows the maximum current per cycle for the middle finger’s MCP and PIP motors, and the wrist joint motor during wrist extension-flexion. Motor current reflects the torque and friction the system must overcome. The stable maximum current over 2.5 hours of continuous operation demonstrates the hand’s robustness, high repeatability, and long-duration capability. Side-mounted fans (\Cref{fig:accuracy}-C) prevent motor overheating, enabling near-continuous use. The experiment was ended voluntarily after 2.5 hours, not due to failure.

To evaluate payload capacity, we used a hand force sensor (electronic grip strength tester) with constant \SI{600}{mA} motor current in position control mode. The hand was tested in various grasps, showing it can hold up to \SI{10.5}{kg} (\SI{103}{N}) using all four fingers, and up to \SI{2}{kg} (\SI{19.6}{N}) using only the index finger.


\subsection{Accuracy and Latency}

\begin{figure}[t!]
\vspace*{0.2cm}

\centering
\includegraphics[width=\linewidth]{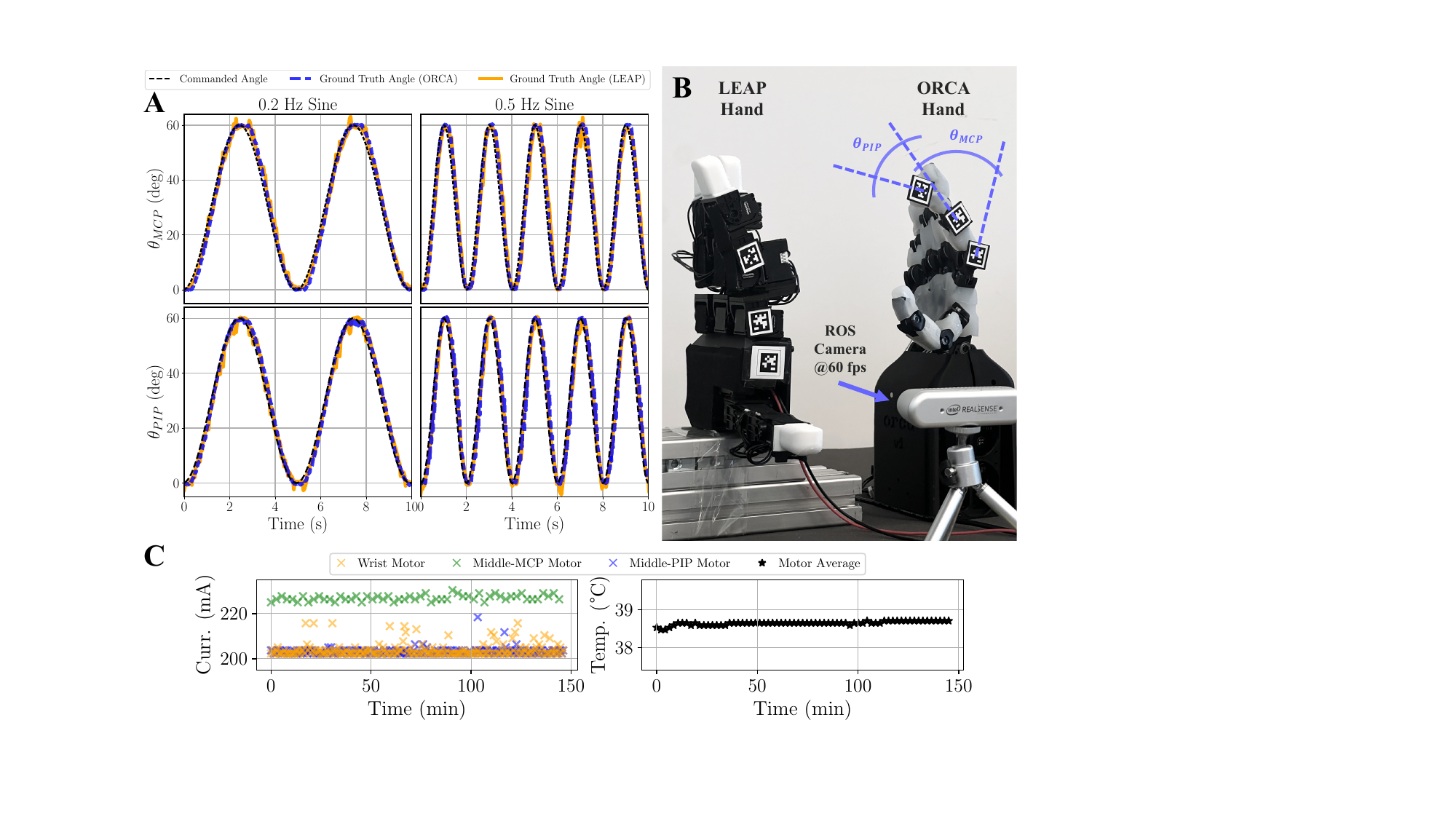}
\caption{(A) Joint response comparison at 0.2 Hz and 0.5 Hz: ORCA achieves accuracy and latency comparable to the LEAP hand despite its tendon-driven design. (B) Accuracy benchmarking setup using AprilTags to infer ground-truth joint angles synchronized with commands. (C) Reliability test: 2.5 hours of continuous grasping (2200+ cycles) and wrist motion (550+ cycles) without failure, overheating, or performance drop, as shown by stable motor current and temperature.}
\label{fig:accuracy}
\vspace{-15pt}
\end{figure}

Accurate control of joint motion is crucial for reliable performance across any dexterous manipulation task. Additionally, excessive latency between action commands and their execution in the real world can severely hinder the implementation of closed-loop behaviors. Incorporating latency into the hand’s model for training in simulation can be helpful to bridge the sim2real gap. To further demonstrate the reliability of the ORCA hand, we, therefore, benchmark the accuracy and latency with which the hand's joints can follow diverse action commands. We propose the following experimental setup to benchmark the accuracy and latency of robotic hands: 

By attaching distinct AprilTags along one finger and measuring their relative orientations to each other, each joint angle on the finger can be calculated with relatively low error (\(\sigma = 0.08^\circ\) for our experimental setup). We place three tag36h11 AprilTags across the MCP and PIP joints of the ORCA hand's index finger and align them in a single plane (Fig.~\ref{fig:accuracy}B). We then position a RealSense D435i camera in front of the AprilTags, using a custom script to ensure that the camera faces the tags at an angle of $90^\circ \pm 5^\circ$ for more precise angle measurements. We record image frames at 60 fps and save them to a ROS2 bag file. Simultaneously, we publish commanded angles that actuate the index finger's MCP and PIP joints in a sine wave pattern and log them into the same ROS2 bag file. This allows us to synchronize the commanded angles with the real angles obtained from the image frames and evaluate the system's latency and accuracy offline.

We compare the accuracy and latency of the ORCA hand with the LEAP \cite{shaw2023leaphandlowcostefficient} hand to benchmark our system's tendon-driven dynamics against the dynamics of a direct-driven robot hand (\Cref{fig:accuracy}-A). We actuate the MCP and PIP joints of both hands' index fingers with 0.2 Hz and 0.5 Hz sine wave patterns. Before each test, we leverage the auto-calibration mechanism of the ORCA hand to account for any changes in tendon length or slack that might have accumulated beforehand. We demonstrate that, through auto-calibration, the finger joints of the ORCA hand accurately follow the commanded sine input. In fact, we achieve similar accuracy to the LEAP hand while being far less bulky thanks to our tendon-driven actuation design. Additionally, we observe that the ORCA hand's joints rotate more smoothly than those of the LEAP hand, which exhibit more jerks, presumably due to the cables and possibly increased inertia interfering with the LEAP hand's finger joint motion. Both hands exhibit average latencies of less than 0.2 seconds, most of which comes from the software process. However, slack in the ORCA hand may introduce additional latency, which is why re-tensioning the spools periodically is important for robust performance.

One limitation of our benchmarking experiment is that we cannot evaluate the accuracy and latency of the hands for faster movements, as the AprilTags can not be tracked at sine wave frequencies above 0.5 Hz or for step signals due to motion blur at 60 fps. Recording the finger joints with a ROS2-compatible camera capable of 240 fps could enable more thorough system identification in the future.

\subsection{Reinforcement Learning}

Reinforcement learning (RL) is commonly used to learn dexterous tasks that are challenging to demonstrate or require fine motor control, such as object reorientation in the hand \cite{handa2024dextremetransferagileinhand}. A key challenge in applying RL to dexterous manipulation tasks is that the policies learned in simulation often perform poorly when transferred to the real robotic hand. We use the IsaacGymEnvs wrapper from \cite{rolling-contact-srl} to train 4096 ORCA hand models in parallel with an advantage actor-critic (A2C) architecture to learn in-hand ball reorientation. We do not use tactile sensors in our RL experiments, primarily due to the additional complexity involved in accurately modeling them. We demonstrate that after one hour of training with domain randomization, we can deploy a robust policy on the physical ORCA hand (\Cref{fig:il_rl}-B), that can successfully reorient a tennis ball along a given rotation axis. 

\subsection{Imitation Learning}
Imitation learning has become another predominant approach in the manipulation community, as it enables learning tasks from a set of demonstrations without requiring task-specific rewards or simulation environments. Various architectures have been proposed to extract meaningful representations of observations and map them to the correct actions \cite{dasari2024ingredientsroboticdiffusiontransformers, chi2024diffusionpolicyvisuomotorpolicy, zhao2023learningfinegrainedbimanualmanipulation}. However, the application of imitation learning to dexterous platforms presents additional challenges mainly due to their higher-dimensional action spaces \cite{qin2022dexmvimitationlearningdexterous, 10769883}.

To demonstrate autonomous task execution with the Orca Hand, we employed a state-of-the-art robotic diffusion transformer \cite{dasari2024ingredientsroboticdiffusiontransformers}. Our setup consists of the hand mounted on a robotic arm (Franka Emika Panda), equipped with two external cameras and one wrist-mounted camera.

Demonstrations were collected using motion capture gloves \cite{rokoko_website}, which provided absolute wrist tracking and finger pose estimation. We retargeted these demonstrations into the robot’s state space using an energy-based minimization objective, similar to \cite{sivakumar2022robotictelekinesislearningrobotic}. The wrist pose was used to control the robotic arm in cartesian end-effector space. This teleoperation method allowed us to showcase the versatility of the hand in a wide range of tasks (\Cref{fig:versatility}) and facilitated the rapid and intuitive collection of demonstrations, even from non-trained operators.


The policy takes as input three camera images along with proprioceptive data from both the robotic arm’s end-effector and the hand. The output consists of an action chunk that predicts future actions.

Over approximately 2h30m, 214 recordings (videos, proprioceptive data, actions) were collected to train the policy on an \textit{NVIDIA GeForce RTX 4090} GPU for 500 epochs, which took approximately 4h. 

\begin{figure}[ht!]
    \vspace*{0.3cm}
    \vspace{-5pt}
    \centering
    \begin{minipage}{0.29\linewidth}
        \centering
        \includegraphics[width=\linewidth]{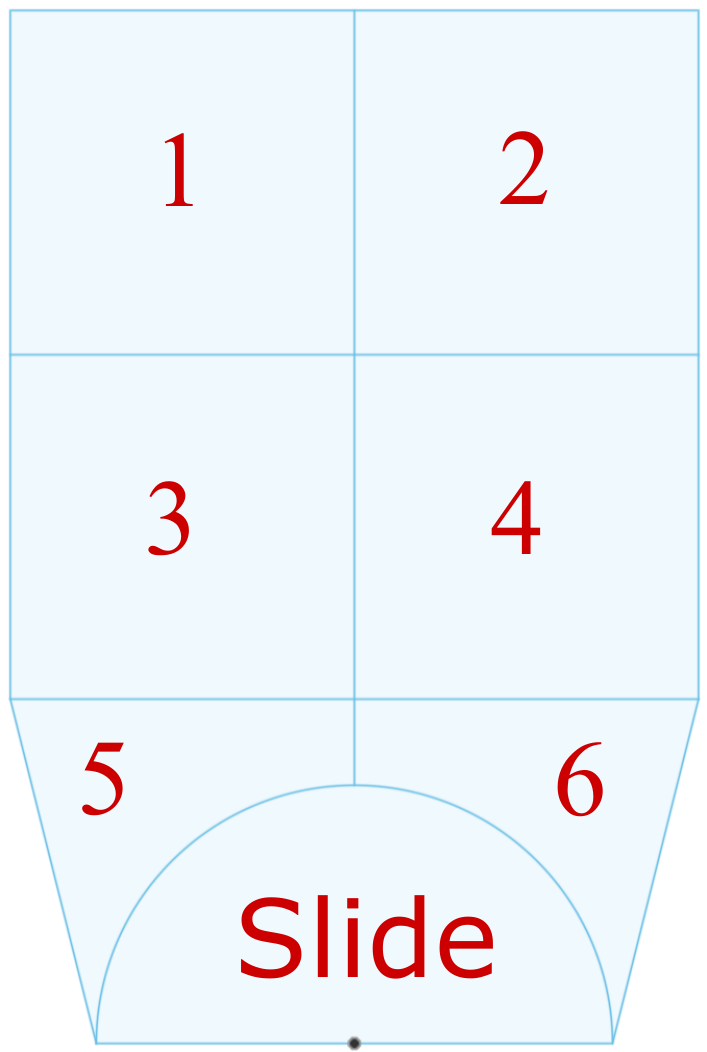}
    \end{minipage}
    \hfill
    \begin{minipage}{0.67\linewidth}
        \begin{tabularx}{\linewidth}{lXXX}
            \toprule
            & \textbf{RGB} & \textbf{Masked} & \textbf{Mixed} \\
            \midrule
            \textbf{Area {\color{red}1}} & 50\% & 80\% & 80\% \\
            \textbf{Area {\color{red}2}} & 10\% & 60\% & 20\% \\
            \textbf{Area {\color{red}3}} & 80\% & 100\% & 100\% \\
            \textbf{Area {\color{red}4}} & 60\% & 100\% & 80\% \\
            \textbf{Area {\color{red}5}} & 60\% & 100\% & 80\% \\
            \textbf{Area {\color{red}6}} & 40\% & 80\% & 60\% \\
            \midrule
            \textbf{Total} & $50$\% & $86.\overset{\_}{6}$\% & $70$\% \\
            \bottomrule
        \end{tabularx}
    \end{minipage}
    \caption{Testing area with respective policy success rates (\% out of 10)}
    \label{tab:policy_success_rates}
    \vspace{-15pt}
\end{figure}

For the proposed task, we performed ablation studies on image pre-processing. Specifically, we compared three policy variations: (1) a baseline policy trained on raw RGB inputs, (2) a policy incorporating segmentation of the cube’s color, and (3) a hybrid approach trained on both data sets (\Cref{tab:policy_success_rates}). 

For (2) and (3), a binary mask was generated using the CIELAB ~\cite{lab_colorspace} color space, with parameters tuned to isolate the red cube. From the RGB image, a three-channel grayscale image is created, and non-zero pixels of the binary mask are set to 255 in the first channel (arbitrarily chosen), effectively highlighting the cube.


\begin{figure}[t!]
\vspace*{0.2cm}
\centering
\includegraphics[width=\linewidth]{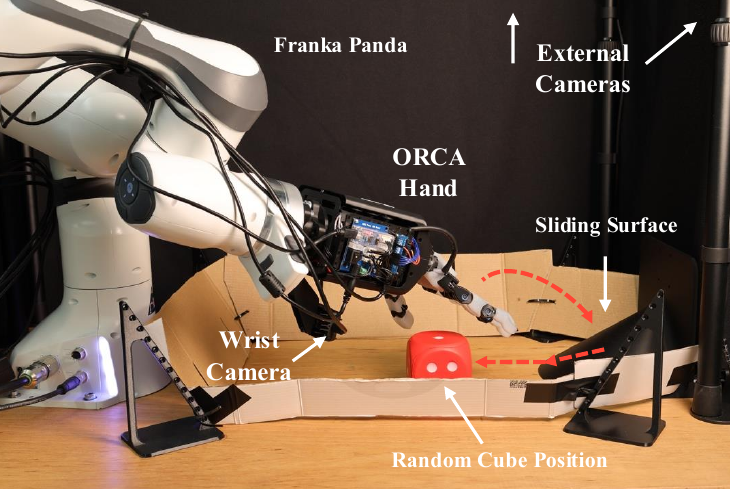}
\caption{Experimental setup for repeated pick \& place: The cardboard serves as a fence, preventing the cube from rolling out of the testing area and enabling uninterrupted, long-duration policy deployment.}
\label{fig:expsetup}
\vspace{-15pt}
\end{figure}

\subsection{Tactile Sensing}

We also evaluated the fingertip sensors. To determine the absolute threshold (AT) of the tactile sensors, a controlled orthogonal force was applied to the front surface of a fingertip using a cylindrical indenter with a diameter of \(2 cm\) and a flat contact surface. The applied force was varied by placing calibrated weights on top of the indenter. The registered touch was classified as any output reading above \(0.01 V\) on the respective analog input on the Arduino. 

\begin{figure*}[t!]
\vspace*{0.2cm}
\centering   
\includegraphics[width=\linewidth]{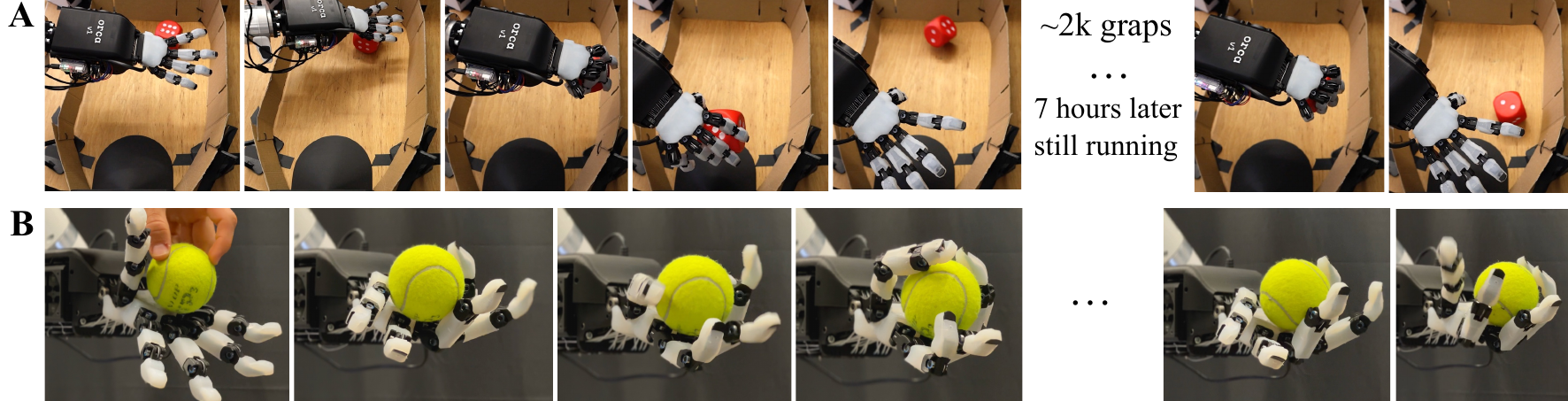}
\caption{(A) Prolonged imitation learning experiments over several hours. Walls and a sliding surface allow for self-resetting of the experiment.
(B) Policies built in simulation are easily deployed zero-shot to the real world since the ORCA hand has autocalibration and only minimal joint control errors.}
\label{fig:il_rl}
\vspace{-10pt}
\end{figure*}

Although the FSR sensors used are rated for a minimum trigger force of \(0.29 N\), the fingertip was able to register forces as low as \(0.05 N\) with perfect accuracy over 10 cycles. It is unknown if this is due to a lower than rated minimum trigger force of the commercial FSR sensors, or if the silicone skin had an unintentional preloading effect on   the sensor. 

However, wear and tear on the silicone skin can drastically affect the AT of the sensors. Although the sensor mounted on the ring finger of the hand showed no degradation after thousands of grasp cycles as part of the experiments described in \Cref{sec:reliability-test}, the AT for the sensor mounted on the pinkie finger increased to \(6.38 N \) due to the degradation of the silicone skin creating an air gap between the silicone skin and the mounted FSR sensor. 

Furthermore, after around 4,500 to 7,000 grasp cycles, the thin copper wires connecting the sensors to the external electronics snapped on the thumb, index, and middle finger. The snapping points occurred at different heights but were all concentrated in the area of the MCP and ABD joints.


\section{Additional Results and Discussion}\label{resultsDiscussion}
In this section, we show the results of human teleoperation and imitation learning with the ORCA hand. We also discuss additional hardware-related findings.

\subsection{Teleoperation - Dexterous Manipulation}
The dexterity and stability of the ORCA hand were successfully evaluated by picking up, interacting with, and placing a variety of objects:
\begin{itemize}
    \item Stack 3 small and large cubes (the same as for IL experiment). The cubes were first placed separately on the table.
    \item Grab a plush toy (about the size of 3 large cubes) from the table top.
    \item Grab a tennis ball lying on the table
    \item Twist open the cap of a \textit{Nutella} jar ($\diameter 8 \, cm$). The jar itself is fully screwed on.
    \item Spin a fidget toy for $2 \, s$. The fidget spinner is placed on a finger by hand, but the grasping and spinning are done purely by teleoperation.
    \item Pick up a pen and write "Hello" on a fastened piece of paper (font size $\sim$200).
    \item Pick up a piece of paper lying on a shut box.
    \item Pick up a cup and pour its contents ($50 \, ml$ water) into another cup.
\end{itemize}

\subsection{Imitation Learning - Repeated Pick \& Place}
To highlight the reliability of both the hardware platform and the trained IL policy, we designed a continuous pick-and-place evaluation task. The robotic hand is required to pick up a cube (6 cm in side length) from a table and place it on a sliding surface, which then causes the cube to fall back onto a random location on the table (\Cref{fig:expsetup}, \Cref{fig:il_rl}-A). 

To evaluate different policies, we collected the ratio of failure positions to success positions (regarding picking up the cube) within a testing area for 60 iterations (10 per sub-area), as shown in \Cref{tab:policy_success_rates}. The most successful policy, using only masked images of the cube, is deployed for 7h\,17min (approx. 2,000 grasping cycles) with no human intervention on the ORCA hand's hardware and minimal intervention in aiding in the pick-and-place task (\Cref{fig:1}-B). Throughout the test, the policy maintained consistent performance, with no tendon slack or rupture, and the experiment was concluded not due to failure but because it sufficiently demonstrated the system’s reliability and effectiveness.
For extended videos and time-lapses of the reliability test, please refer to the \href{https://www.orcahand.com}{ORCA project website}.

\subsection{Tactile Sensing}
While integrated tactile sensing provides a cost-effective and low barrier-of-entry approach to providing tactile feedback at the fingertips, the current design still shows a variety of limitations regarding reliability over thousands of grasp cycles. These limitations are namely the degradation of the silicone skin that is vital to reliably transmit the contact forces to the sensors (occurrence in 2 fingertips after approx. 2,000 to 4,000 grasp cycles) as well as the snapping of the thin copper wires used for signal transmission (occurrence in 3 fingertips after approx. 4,500 to 7,000 grasp cycles). Both limitations will be addressed in future work to allow for reliable tactile sensing integration into autonomous tasks.




\section{Conclusion}\label{conclusion}

The ORCA hand provides an accessible platform for advancing robotic manipulation, aiming to support real-world tasks and cutting-edge research. It is designed to be human-like, compliant, robust, versatile, easy to control, and cost-effective.

Despite these strengths, limitations remain. Prolonged use requires manual re-tensioning to maintain performance. To overcome this, we plan to develop an autonomous re-tensioning mechanism to reduce tendon slack without human intervention, enabling longer operation and supporting real-world RL applications. Future work also includes integrating sensors directly into the learning pipeline and applying more advanced deep learning methods to handle complex environmental interactions.

\addtolength{\textheight}{-1.1cm}   

\section*{ACKNOWLEDGMENT}
We thank Robert Jomar Malate and the \href{https://rwr.ethz.ch}{Real World Robotics course} team at ETH Zurich for initiating and enabling this project, and Carlo Sferrazza and the Berkeley Robot Learning Lab for access to their LEAP Hand. This work was funded through a research collaboration with armasuisse. Filippos Katsimalis acknowledges support from the Bodossaki Foundation, SYN-ENOSIS, and the Onassis Foundation. Aristotelis Sympetheros acknowledges support from the John S. Latsis Public Benefit Foundation and SYN-ENOSIS.

\bibliographystyle{IEEEtran}
\bibliography{references}

\end{document}